\documentclass[10pt,journal,compsoc]{IEEEtran}
\ifCLASSOPTIONcompsoc
  \usepackage[nocompress]{cite}
\else
  \usepackage{cite}
\fi
\usepackage{latexsym,mathrsfs}
\usepackage{amsmath,amssymb} 
\usepackage{amsthm,enumerate,verbatim}
\usepackage{amsfonts}
\usepackage{graphicx}
\usepackage{algorithm}
\usepackage{algorithmic}
\usepackage{url,color}
\usepackage{subcaption}

\DeclareMathOperator*{\argmax}{arg\,max}
\DeclareMathOperator*{\argmin}{arg\,min}

\newcommand{\black}[1]{\textcolor{black}{#1}} 

\newtheorem{proposition}{Proposition}
\newtheorem{remark}{Remark}

\definecolor{brightpink}{rgb}{1.0, 0.0, 0.5}

\newcommand{\ngi}[1]{{{\color{black} #1}}}

\newcommand{\hien}[1]{\textcolor{black}{#1}} 
\newcommand{\revise}[1]{\textcolor{black}{#1}} 
\newcommand{\red}[1]{{\color{black}#1}}
 \newcommand{\rd}[1]{{\color{black}#1}}
 
 \newcommand{\reviselast}[1]{\textcolor{black}{#1}} 
 
 \newcommand{\reviselastlast}[1]{\textcolor{black}{#1}}

\ifCLASSINFOpdf
\else
\fi

\begin{document}
%
\title{Distributionally Robust and Multi-Objective  \\ 
Nonnegative Matrix Factorization}
%
%
%
%

\author{Nicolas~Gillis,~\IEEEmembership{Member,~IEEE,}
Le~Thi~Khanh~Hien,~Valentin~Leplat,~\\  
and~Vincent~Y.~F.~Tan,~\IEEEmembership{Senior Member,~IEEE} 
\IEEEcompsocitemizethanks{\IEEEcompsocthanksitem N.\@ Gillis, L.\@ T.\@ K.\@ Hien and V. Leplat are with Department of Mathematics and Operational Research, 
Facult\'e Polytechnique, Universit\'e de Mons,  
Rue de Houdain 9, 7000 Mons, Belgium. This work was supported by the Fonds de la Recherche Scientifique - FNRS and the Fonds Wetenschappelijk Onderzoek - Vlanderen (FWO) under EOS Project no O005318F-RG47, and by the European Research Council (ERC starting grant no 679515). \protect\\
E-mails: \{nicolas.gillis, thikhanhhien.le, valentin.leplat\}@umons.ac.be 
\IEEEcompsocthanksitem V.\@ Y.\@ F.\@ Tan is with the Department of Electrical and Computer Engineering, 
Department of Mathematics,  
National University of Singapore,  
Singapore 119077. This work is also supported by a Singapore National Research Foundation (NRF) Fellowship (R-263-000-D02-281). \protect\\
E-mail:  vtan@nus.edu.sg 
}
\thanks{Manuscript accepted, February 2021.}
}

%
%

\markboth{IEEE Trans. Pattern Anal. Mach. Intell.,~Vol.~X, No.~Y, Month~Year}%
{Gillis \MakeLowercase{\textit{et al.}}: Distributionally Robust and Multi-Objective NMF}
%



\IEEEtitleabstractindextext{%
\begin{abstract}
Nonnegative matrix factorization (NMF) is a linear dimensionality reduction technique for analyzing nonnegative data. A key aspect of NMF is the choice of the objective function that depends on the noise model (or statistics of the noise) assumed on the data. In many applications, the noise model is unknown and difficult to estimate. 
In this paper, we define a multi-objective NMF (MO-NMF) problem, where several objectives are combined within the same NMF model. 
We propose to use Lagrange duality to judiciously optimize for a set of weights to be used within the framework of the weighted-sum approach, that is, we minimize a single objective function which is a weighted sum of the all objective functions. 
We design a simple algorithm based  on multiplicative updates to minimize this weighted sum. 
We show how this can be used to find distributionally robust NMF (DR-NMF) solutions, that is, solutions that minimize the largest error among all objectives, \color{black} using a dual approach solved via a heuristic inspired from the Frank-Wolfe algorithm. \color{black} 
We illustrate the effectiveness of this approach on synthetic, document and audio data sets. The results show that DR-NMF is robust to our incognizance of the noise model of the NMF problem. 
\end{abstract}

}

\maketitle

\IEEEdisplaynontitleabstractindextext

%
\IEEEpeerreviewmaketitle

\IEEEraisesectionheading{\section{Introduction}}

Nonnegative matrix factorization (NMF) consists in the following problem: Given a nonnegative matrix $X \in \mathbb{R}^{m \times n}_+$ and a factorization positive rank $r \ll \min(m,n)$, find two nonnegative matrices $W \in \mathbb{R}^{m \times r}_+$ and $H \in \mathbb{R}^{r \times n}_+$ such that $WH \approx X$. 
NMF is a linear dimensionality reduction technique for nonnegative data. In fact, assuming each column of $X$ is a data point, it is reconstructed via a linear combination of $r$ basis elements given by the columns of $W$ while the columns of $H$ provide the weights (or coefficients) to reconstruct each column of $X$ within that basis, that is, for all $j$, 
\[
X(:,j)  \approx \sum_{k=1}^r W(:,k) H(k,j). 
\]
NMF has attracted a lot of attention since the seminal paper of Lee and Seung~\cite{lee1999learning}, with applications in image analysis, document classification and music analysis. See for example \cite{cichocki2006new, gillis2014} and the references therein. 
Many NMF models have been proposed over the years. They mostly differ in  two aspects: 
\begin{enumerate}
\item {\em Additional constraints} are added to the factor matrices $W$ and $H$ such as sparsityÌ~\cite{hoyer2004non}, 
spatial coherence~\cite{liu2011approach} or 
smoothness~\cite{essid2013smooth}. 
These constraints are motivated by a priori information on the sought solution and depend on the application at hand. 
Note that these additional constraints are in most cases imposed via a penalty term in the objective function. 

\item The choice of the {\em objective function} that assesses the quality of an approximation by evaluating some distance between $WH$ and $X$ differs. This choice is usually motivated by the noise model/statistics assumed on the data matrix $X$.  
The most widely used class of objective functions are component-wise and based on the $\beta$-divergences defined as follows: for $x,y \in \mathbb{R}_+$, 
\begin{align*}
&D_{\beta}(x,y) \\
 &=  \left\{ 
\begin{array}{cc}
  \frac{x}{y} - \log \frac{x}{y} - 1  & \text{for }  \beta = 0, \\
 x \log \frac{x}{y} -x + y & \text{for } \beta = 1, \\ 
\frac{1}{\beta (\beta-1)} \left(x^\beta + (\beta-1)y^\beta - \beta xy^{\beta-1}\right) &  \text{for } \beta \neq 0,1. 
 \end{array}
\right. 
\end{align*}
We will use the following matrix-wise notation,  
\[
D_{\beta}(X,WH) = \sum_{i,j} D_{\beta}\left(X_{ij},(WH)_{ij} \right). 
\] 
\revise{Minimizing the $\beta$-divergence in NMF is equivalent to maximizing the log-likelihood of the NMF model under different noise distributions~\cite{fevotte2011algorithms,tan2013automatic}.} The following special cases are of particular interest (see for example~\cite{fevotte2011algorithms} for a discussion): 
\begin{itemize}
\item $D_{2}(X,WH) = \frac{1}{2}\|X-WH\|_F^2$ is the Frobenius norm (additive Gaussian noise). 
\item $D_{1}(X,WH) = \text{KL}(X,WH)$  is the Kullback-Leibler (KL) divergence (Poisson noise). 
\item $D_{0}(X,WH) = \text{IS}(X,WH)$ is the Itakura-Saito (IS) divergence (multiplicative Gamma noise). 
\end{itemize}
\end{enumerate} 

In this paper, we focus on the second aspect, namely, the choice of the objective function. We will consider a multi-objective NMF (MO-NMF) formulation.  {More precisely}, we will consider a weighted sum of the different objective functions, which is arguably one of the most widely used approach in multi-objective optimization~\cite{marler2010}.  
Our main motivation to consider this class of models is that in many applications it is not clear which objective function to use because the statistics of the noise is unknown. 
To the best of our knowledge, there are currently   three main classes of methods to handle this situation: 
\begin{itemize}

\item The user chooses  the objective function she/he believes is the most suitable for the application at hand. This is, as far as we know, the simplest and most widely-used approach. However, this approach is an {\em ad hoc} one.

\item The objective function is automatically selected using cross-validation, where the training is done on a subset of the entries of the input data matrix and the testing on the remaining entries~\cite{mollah2007robust,choi2010learning}. 

\item The most suitable objective function is chosen using some statistically motivated criteria such as 
score matching~\cite{lu2012selecting} or 
maximum likelihood~\cite{dikmen2015learning}. 

\end{itemize}

However, in all the above approaches, 
if the choice of the objective function is wrong, 
the NMF solution provided could be far from the desired solution (as we will show in our numerical experiments in Section~\ref{sec:ne}). 
Another possibility which we propose in this paper is to compute an NMF solution that is robust to different types of noise distributions; this is referred to as \emph{distributionally robust}, and is closely related to robust optimization~\cite{ben2009robust}. 
In mathematical terms, we will consider the problem 
\begin{equation*}
\min_{(W,H) \geq 0} \max_{\beta \in \Omega} D_{\beta}(X,WH), 
\end{equation*}
As we will see, this problem can be tackled by minimizing a weighted sum of the different objective functions~\cite{marler2010}, exactly as for MO-NMF, but where the weights assigned to the different objective functions are {\em automatically tuned} within the iterative process.

\paragraph*{Outline of the paper} 
 
 In Section~\ref{sec:monmf}, we first define MO-NMF and explain how to scale the objective functions to make the comparison between the constituent NMF objective functions.   
 Then we give our main motivation to consider MO-NMF, namely to be able to compute distributionally robust NMF (DR-NMF) solutions, that is, solutions that minimize the largest objective function value. 
 In Section~\ref{sec:algo}, we propose simple multiplicative updates (MU)  \textcolor{black}{to solve a weighted-sum approach for MO-NMF. In Section~\ref{sec:algoDRNMF}, we propose a heuristic scheme to solve DR-NMF which  updates the primal variables using the MU  
 and the dual variable using the Frank-Wolfe descent direction. } 
 Finally, we illustrate in Section~\ref{sec:ne} the effectiveness of our approach on synthetic, document and audio data sets.

\section{Multi-Objective NMF (MO-NMF)} \label{sec:monmf}

Let $\Omega$ be a finite subset of $\mathbb{R}_+$. 
We consider in this paper the following MO-NMF problem: 
\[
\min_{(W, H) \geq 0} \{ D_{\beta}(X,WH) \}_{\beta \in \Omega} . 
\] 
Note that we focus on $\beta$-divergences to simplify our presentation and because these are the most widely-used divergences to measure the ``distance'' between the given matrix $X$ and its approximation $WH$ in the NMF literature.  However, our approach can adapted to be used for other objectives functions (for example, $\alpha$-divergences~\cite{cichocki2009nonnegative}). 
To tackle this problem, we consider the standard weighted-sum approach~\cite{deb2014multi} which consists in solving the following minimization problem which involves a {\em single} objective function:  
\begin{equation*}
\min_{(W,H)  \geq 0} D_{\Omega}^\lambda(X,WH), 
\end{equation*} 
where $D_{\Omega}^\lambda(X,WH)
= 
\sum_{\beta \in \Omega} \lambda_{\beta} D_{\beta}(X,WH)$,  
 $\lambda \in \mathbb{R}^{|\Omega|}_+$, and 
$\|\lambda\|_1=\sum_{\beta \in \Omega} \lambda_\beta = 1$. 
Using different values for $\lambda$ allows to generate different Pareto-optimal solutions. See Section~\ref{sec:pareto} for some examples. Note, however, that it does not allow to generate all Pareto-optimal solutions~\cite{deb2014multi}. 
A Pareto-optimal solution is a solution that is not dominated by any other solution. That is, $(W,H)$ is a Pareto-optimal solution if there does not exist a feasible solution $(W',H')$ such that 
\begin{itemize} 
\item $D_{\beta}(X,W'H') \leq D_{\beta}(X,WH)$ for all  $\beta \in \Omega$, and  

\item there exists $\beta \in \Omega$  such that  $D_{\beta}(X,W'H') < D_{\beta}(X,WH)$. 
\end{itemize}

Multi-objective optimization has already been considered for NMF problems. 
However, 
most of the existing literature considers combining a single data fitting term with penalty terms on the factor matrices; for example, an $\ell_1$ penalty to obtain sparse solutions~\cite{gong2018multiobjective}. 
As far as we know, 
the only paper where several objectives are used to balance  different data fitting terms is~\cite{zhu2016biobjective}. 
The authors combined two objectives, one being a standard data fitting term (more precisely, they used the Frobenius norm $\|X-WH\|_F^2$) 
and the other being a data fitting term in a feature space obtained using a nonlinear kernel (that is, a term of the form $\|\Phi(X)-\Phi(W)H\|_\mathcal{H}^2$ where $\|.\|_\mathcal{H}$ corresponds to the norm in the feature space). 
Hence this approach is rather different than ours where we allow more than two objectives and where we only focus on the input space. Moreover, we will optimize the weights in a principled optimization-theoretic fashion, whereas~\cite{zhu2016biobjective} uses an {\em ad hoc} manner to combine the two terms. 
Another related work~\cite{csimcsekli2015learning} considers a data fusion problem where several data sets, denoted $X_1, X_2, \dots, X_p$, share the same factor $H$. 
Their goal is to compute $H \geq 0$ and $W_i \geq 0$ such that 
$X_i \approx W_i H$ for $i=1, 2, \dots, p$. To achieve this goal, the authors use a weighted objective function 
$\sum_{i=1}^p \lambda_i D_{\beta_i}(X_i, WH_i)$ for some well-chosen weights $\lambda_i$'s, and some parameters $\beta_i$'s that depend on the noise statistic of the corresponding data set. Again, this is a rather different setup that ours as there is no distributionally robust aspect. 

\subsection{Scaling of the objectives}   \label{sec:scaling}

It can be easily checked that for any constant $\alpha > 0$, we have 
\[
D_{\beta}(\alpha X,\alpha WH) = \alpha^\beta D_{\beta}(X,WH). 
\]
Hence the values of the divergences for different values of $\beta$ depend highly on the scaling of the input matrix. 
This is usually not a desirable property in practice, since most data sets are not particularly properly scaled and since scaling simply multiplies the noise by a constant which in most cases does not change its distribution (only its parameters). 
Therefore, we will scale the objectives to have a meaningful linear combination, in the sense that each term in the sum has the same importance. It will be particularly crucial for our DR-NMF model described in the next section. In fact, as  we will see in Section~\ref{sec:ne}, DR-NMF will generate solutions that have small error for all objectives
instead of just one; and as such, 
the solutions inherit superior qualities of the ones generated by different divergences.
We will use the following approach to scale the different objective functions. 
 First, we compute a solution $(W_{\beta}, H_{\beta})$ for 
 $\min_{(W,H) \geq 0}D_\beta(X,WH)$ to obtain the error $e_\beta = D_{\beta}(X, W_{\beta} H_{\beta})$. 
Note that we can only compute this minimization in an approximate fashion because the NMF problem is NP-hard~\cite{vavasis2009complexity}. 
Then, we define 
\[
\bar{D}_{\beta}(X, W H) 
= 
\frac{D_{\beta}(X, W H)}{e_\beta}, 
\]
so that $\bar{D}_{\beta}(X, W_{\beta} H_{\beta})  = 1$. 
Finally, we will only consider the MO-NMF problem where the objectives ${D}_{\beta}(X, W H)$ are replaced by their normalized versions $\bar{D}_{\beta}(X, W H)$, that is, 
\begin{equation}
\min_{(W,H)  \geq 0} \bar{D}_{\Omega}^{\lambda} (X,WH), \label{wsnmf}
\end{equation}
where $\bar{D}_{\Omega}^{\lambda} (X,WH) 
= \sum_{\beta \in \Omega} \lambda_{\beta} \bar{D}_{\beta}(X,WH)$.   
 In Section~\ref{sec:algo}, we propose a MU algorithm to tackle this problem.

\subsection{Main motivation: Distributionally robust NMF}

If the noise model on the data is unknown, but it is known that it corresponds to a distribution associated with a $\beta$-divergence with $\beta \in \Omega$ (for example, the Tweedie distribution as discussed in \cite{tan2013automatic}),  it makes sense to consider the following distributionally robust NMF (DR-NMF) problem 
\begin{equation} \label{drnmf}
\min_{(W,H) \geq 0} \max_{\beta \in\Omega} \bar{D}_{\beta}(X,WH). 
\end{equation}
We use $\bar{D}_{\beta}(\cdot ,\cdot )$, not $D_\beta(\cdot,\cdot)$,  because otherwise, in most cases, the above problem amounts to minimizing a single objective corresponding to the $\beta$-divergence with the largest value; 
see the discussion in Section~\ref{sec:scaling} where $\Omega$ is a subset of $\beta$'s of interest. 
In Section~\ref{sec:algoDRNMF}, we will design an algorithm to tackle this problem based on MO-NMF. 
 We remark that, as mentioned in the introduction, 
Problem~\eqref{drnmf} is intrinsically a deterministic robust optimization problem. However, since each $\beta$-divergence is associated to a distribution of noise (see some examples in the introduction), we prefer using the name DR-NMF for Problem~\eqref{drnmf} to emphasize its essence, which is finding a solution that is robust to different types of noise distributions.

\section{Multiplicative updates for \eqref{wsnmf}} 
\label{sec:algo}

In this section, we propose MU for~\eqref{wsnmf} which we will be able to use \rd{as a subroutine} to tackle MO-NMF and DR-NMF. 
As with most NMF algorithms, we use an alternating strategy; that is, we will first optimize  over the variable $W$ for fixed $H$ and then reverse their roles. 
By the symmetry of the problem ($X^T = H^T W^T$), we will focus on the update of $H$; the update of $W$ can be obtained similarly.

\subsection{Deriving MU} 

Let us recall the standard way MU are derived (see for example  \cite{lee2001algorithms, fevotte2011algorithms, yang2011unified}) on the following general optimization problem with nonnegativity constraints 
\begin{equation} \label{nnopt}
\min_{x \geq 0} f(x).  
\end{equation}
Let us apply a rescaled gradient descent method to~\eqref{nnopt}, that is, use the following update 
\[
x^+ = x - B \nabla f(x), 
\]
where $x$ is the current iterate, $x^+$ is the next iterate, and $B$ is a diagonal matrix with positive diagonal elements.  
Let $\nabla_{+} f(x) > 0$ and 
$\nabla_{-} f(x) > 0$ be such that 
$\nabla f(x) = \nabla_{+} f(x) - \nabla_{-} f(x)$. 
Taking $B_{ii} = \frac{x_i}{\nabla_{+} f(x)_i}$ for all $i$, 
we obtain the following MU  rule:
\begin{align}
x^+ 
& = x - \frac{[x]}{[\nabla_{+} f(x)]} \circ \left( \nabla_{+} f(x) - \nabla_{-} f(x)\right) \nonumber \\
&= x \circ \frac{[\nabla_{-} f(x)]}{[\nabla_{+} f(x)]},  \label{muderiv}
\end{align} 
where $\circ$ (resp.\@ $[\cdot]/[\cdot]$) refers to component-wise multiplication (resp.\@ division) between two vectors or matrices. 
Note that we need strict positivity of $\nabla_{+} f(x)$ and 
$\nabla_{-} f(x)$, otherwise we would encounter problems involving division by zero or a variable directly set to zero, which is not desirable. 
Using the above simple rule with proper choices for $\nabla_{+} f(x)$ and $\nabla_- f(x)$ leads to algorithms that are, in many cases, guaranteed to not increase the objective function, that is, $f(x^+) \leq f(x)$; see below for some examples, and~\cite{yang2011unified} for a discussion and an unified rule to design such updates. 
This is a desirable property since it avoids any line-search procedure and also preserves non-negativity naturally. 
If we cannot guarantee that the updates are non-increasing, the step length can be reduced, that is, use 
\[
x^+_\gamma = x - \gamma B \nabla f(x), 
\]
for some $0 < \gamma \leq 1$ which leads to 
\[
x^+_\gamma = (1-\gamma) x + \gamma x^+ . 
\]
For example, one can set the step size $\gamma = 1/2^k$ for the smallest $k$ such that the error decreases; such a $k$ is guaranteed to exist since the rescaled gradient direction is a descent direction. We implemented such a line search; see Algorithm~\ref{muHwsnmf} below. 
This idea is similar to that in~\cite{lin2007convergence}.  
Moreover, it would be worth investigating the use of regularizers to guarantee convergence to stationary points without the use of a line search~\cite{zhao2018unified}.  
 
For $x_i=0$, we have that $B_{ii} = 0$ and the MU are not able to modify $x_i$: this is the so-called zero-locking phenomenon~\cite{berry2007algorithms}. 
A possible way to fix this issue in practice is to use a lower bound $\epsilon$ on the entries of $x$, say  $\epsilon=10^{-16}$, replacing $x^+$ with $\max(\epsilon,x^+)$. 
This allows such algorithms to be guaranteed to converge to a stationary point of $\min_{x \geq \epsilon} f(x)$~\cite{gillis2011nonnegative, takahashi2014global}. More precisely, any sequence of solutions generated by the modified MU has at least one convergent subsequence and the limit of any convergent subsequence is a stationary point~\cite{takahashi2014global}.  Moreover, it can also be shown~\cite[Chap.~4.1]{gillis2011nonnegative} that such stationary points are close to stationary points of the original problem in~\eqref{nnopt}. 
We will use this simple strategy in this paper.

\subsection{Multiplicative Updates  for~\eqref{wsnmf}} 

We now provide more details on how to choose $\nabla_- f(x)$ and $\nabla_+ f(x)$ for the family of $\beta$-divergences in order to tackle~\eqref{wsnmf}. For all $\beta$, we have 
\[
\nabla^H D_{\beta}(X,WH)
= 
\nabla^H_+ {D}_{\beta}(X,WH) - \nabla^H_- {D}_{\beta}(X,WH), 
\]
where $\nabla^H$ denotes the gradient with respect to variable $H$, and 
\begin{align*}
&\nabla^H_+ {D}_{\beta}(X,WH) = 
W^T (WH)^{\circ (\beta-1)} 
 \quad \text{ and } \\
&\nabla^H_- {D}_{\beta}(X,WH) = 
W^T \left( (WH)^{\circ (\beta-2)} \circ X \right), 
\end{align*} 
where  $A^{\circ k}$ is the component-wise exponentiation by $k$ of the matrix $A$. 
To solve~\eqref{wsnmf} using MU, we simply use the linear combination of the above standard choice~\cite{tan2013automatic}; see  
Algorithm~\ref{muHwsnmf} for the update of $H$ (the update for $W$ is obtained in the same way by symmetry). Note that the line-search procedure (steps 3 to 6) is very rarely entered (we have only observed it in all our numerical experiments described in Section~\ref{sec:ne} when $\Omega = \{0\}$, that is, only for IS-NMF alone). Note also that the only difference between  $\bar{D}_{\beta}$ and ${D}_{\beta}$ is a constant term; see Section~\ref{sec:scaling}. 
\rd{In the case of a single} objective \rd{(i.e., that $|\Omega|=1$)}, Algorithm~\ref{muHwsnmf} particularizes to the standard MU algorithm for NMF; 
see for example~\cite{fevotte2011algorithms} and the references therein.

\algsetup{indent=2em}
\begin{algorithm}[ht!]
\caption{MU for $H$ to solve~\eqref{wsnmf}} \label{muHwsnmf}
\begin{algorithmic}[1]
\REQUIRE
The matrices $X \in \mathbb{R}^{m \times n}_+$ and 
$W \in \mathbb{R}^{m \times r}_+$, 
an initialization $H \in \mathbb{R}^{m \times r}_+$, 
a finite set $\Omega$ of nonnegative real numbers, and 
$\lambda \in \mathbb{R}^{|\Omega|}_+$. 

\ENSURE $H^+_\gamma \in \mathbb{R}^{m \times r}_+$ such that 
$\bar{D}_{\Omega}^\lambda(X,WH^+_\gamma) 
\leq 
\bar{D}_{\Omega}^\lambda(X,WH)$. 
\medskip

\STATE $H^+ = H \circ \frac{ \left[ \sum_{\beta \in \Omega} \lambda_\beta
\big( \nabla^H_-  \bar{D}_{\beta}(X,WH) \big) \right] 
}{
\left[ \sum_{\beta \in \Omega} \lambda_\beta
\big( \nabla^H_+  \bar{D}_{\beta}(X,WH) \big) \right]}$. 

\STATE  $\gamma = 1$, $H^+_1 = H^+$. 

\WHILE { $\bar{D}_{\Omega}^\lambda(X,WH^+_\gamma) 
>  
\bar{D}_{\Omega}^\lambda(X,WH)$ }

\STATE $\gamma = \frac{\gamma}{2}$. 

\STATE  $H^+_\gamma = (1-\gamma) H + \gamma H^+$. 

\ENDWHILE  

\end{algorithmic}
\end{algorithm} 

Because of the step length procedure that guarantees the objective function to not increase (steps 3-6), 
the use of Algorithm~\ref{muHwsnmf} in an alternating scheme to solve~\eqref{wsnmf} by updating $W$ and $H$ alternatively is guaranteed to not increase  the objective function. Since the objective function is bounded below, this guarantees that the objective function values 
converge as $k$ goes to infinity. 

\section{Algorithm for DR-NMF} \label{sec:algoDRNMF}

{\color{black} As $(W,H) \mapsto \max_{\beta \in \Omega} \bar{D}_{\beta}(X,WH)$ is a non-convex function, obtaining a global solution $(W^*,H^*)$ for~\eqref{drnmf} \black{efficiently} is not possible in general. 
In particular, deciding whether the minimum in~\eqref{drnmf} is equal to zero (that is, deciding whether there exists $W$ and $H$ such that $X = WH$) is NP-hard~\cite{vavasis2009complexity}.    
In the following, we propose to find an approximate  solution for the DR-NMF problem via a weighted sum of the different objective functions. }
 We first observe that $$\max_{\beta \in \Omega} \bar{D}_{\beta}(X,WH)=\max_{\lambda\geq 0, \sum_{\beta \in\Omega} \lambda_\beta=1}\sum_{\beta \in\Omega} \lambda_\beta \bar D_\beta (X,WH).$$
Hence \eqref{drnmf} can be reformulated as 
\begin{equation}
\label{primal}
 \min_{(W,H) \geq 0} \max_{\lambda\geq 0, \| \lambda\|_1 = 1}\sum_{\beta \in\Omega} \lambda_\beta \bar D_\beta (X,WH).  
\end{equation}

\subsection{Related works on min-max problems} 
\label{sec:related} 
{\color{black}
\black{The problem in~\eqref{primal} is a min-max problem.} 
Let us \black{present} a brief review of well-known methods for solving a \black{general min-max problem, also known as a {\em saddle point problem} (SSP), of the form}
\begin{equation}
\label{eq:saddlepointPhi}
\min_{x \in \mathcal X} \max_{y\in \mathcal Y} \Phi(x,y),
\end{equation} where $ \mathcal X$ and $\mathcal Y$ are closed convex sets. \black{SPPs are} abound in game theory, machine learning and statistics. \black{A special class of SPPs is the class of
bilinear SPPs which assume that the objective can be expressed as} $\Phi(x,y)=f(x)+\langle Ax,y \rangle + g(y)$ where $A$ is a linear operator, $f$ and $g$ are differentiable functions, and the coupling between $x$ and $y$ is linear in $x$ and linear in $y$. 
Bilinear SPPs  \black{have} been extensively studied and can be solved efficiently by several methods such as Nesterov's smoothing method~\cite{Nest_05} and the primal-dual hybrid gradient method~\cite{Chambolle_16, Chen_14}. For non-bilinear SPPs, \black{the} proximal mirror descent method (which subsumes \black{the} proximal gradient descent method as a special case~\cite{Nemi_05,Juditsky_12b,mertikopoulos2018optimistic}),  is often the method of choice since it is a direct method applied to the underlying SPP (while \cite{Nest_05} requires advanced smoothing techniques) and it can be \black{adapted to the case in which regularizers or constraints are present, assuming   the involving proximal maps can be computed}. In another line of works, the approach of sequentially solving auxiliary sub-problems (which can have closed-form solutions or can be approximately solved by suitable solvers) 
to alternatively update $x$ (while fixing $y$) and $y$ (while fixing~$x$) \black{has also been} developed in the literature~\cite{Hamedani2018,Hien2019,Lu2019}.  

To establish convergence \black{guarantees} of \black{algorithms for SPPs}, the following three typical assumptions are made in the literature: 
(A)~$\Phi$ is convex in $x$ and concave in $y$, 
(B)~the gradient map $(x,y)\mapsto [\nabla_x \Phi(x,y),-\nabla_y \Phi(x,y)]$ is Lipschitz continuous, and 
(C)~the SPP has at least one saddle-point, that is, there exists $(x^*,y^*)$ such that 
\begin{equation}
\label{defsaddlepoint}
\Phi(x^*,y)\leq \Phi(x^*,y^*)\leq \Phi(x,y^*) \text{ for all } x\in \mathcal X \text{ and } y\in \mathcal Y.
\end{equation} 
 Assumption (C) is satisfied when $\mathcal X$ and $\mathcal Y$ are convex compact sets and $\Phi$ is a continuous  convex-concave function \cite[Proposition 5.5.3]{Bertsekas2009}, hence \hien{Assumption (C)} can be omitted for SPPs under these settings~\cite{Nemi_05,Juditsky_12b}. For SPPs under other settings 
 (for example when $\mathcal X$ or $\mathcal Y$ is unbounded, or when $\Phi(x,y)$ is not convex-concave), Assumption (C) \red{concerning} the existence of saddle points is a standard \black{assumption for the development of} numerical algorithms for solving SPPs; see~\cite{Chen_14,Hamedani2018,mertikopoulos2018optimistic} and references therein. Our SPP \eqref{primal} \red{neither satisfies}   Assumption (A) nor Assumption (B) since $(W,H)\mapsto \bar{D}_\beta(X,WH)$ is not convex and $(W,H)\mapsto \nabla \bar{D}_\beta(X,WH)$ is not Lipschitz continuous. 
\black{These facts prevents} us  \black{from applying standard} SPP algorithms with convergence guarantees \black{to solve}~\eqref{primal}. 


\subsection{Preliminaries}


The following proposition provides some \black{properties} of saddle points of $\Phi$. Its proof can be derived from \cite[Proposition 3.4.1]{Bertsekas2009} and  the definition of subgradients.
\begin{proposition}
\label{prop:SPP}
Consider the SPP in~\eqref{eq:saddlepointPhi}. Suppose that \red{for each $(x,y) \in {\cal X}\times{\cal Y}$},  $\Phi(\cdot,y) $ and   $-\Phi(x,\cdot)$ are \black{subdifferentiable} \red{on ${\cal X}$ and ${\cal Y}$ respectively}. 
Then, 

\noindent (I) $(x^*,y^*)$ is a saddle point of \eqref{eq:saddlepointPhi} if and only if there exist \black{a} subgradient $\Phi'_x(x^*,y^*)$ of $\Phi(\cdot,y^*)$ at $x^*$ and \black{a} subgradient $-\Phi'_y(x^*,y^*)$ of $-\Phi(x^*,\cdot)$ at $y^*$ such that 
$$x^*=\Pi_{\mathcal X}(x^*-\Phi'_x(x^*,y^*))\;\;  \text{and}\;\; y^*=\Pi_{\mathcal Y}(y^*+\Phi'_y(x^*,y^*)).
$$

\noindent (II) $(x^*,y^*)$ is a saddle point of \eqref{eq:saddlepointPhi} if and only if   strong duality holds, that is, 
$$\min_{x\in \mathcal X} \max_{y \in \mathcal Y} \Phi(x,y)= \max_{y \in \mathcal Y} \min_{x\in \mathcal X} \Phi(x,y), 
$$
and $$x^*\in \argmin_{x\in \mathcal X} \max_{y \in \mathcal Y} \Phi(x,y),\quad y^*\in \argmax_{y \in \mathcal Y} \min_{x\in \mathcal X} \Phi(x,y).$$
\end{proposition}
Suppose $\Phi$ has a saddle point. Proposition \ref{prop:SPP} shows that if we can find a solution of the dual problem  $y^* \in \argmax_{y \in \mathcal Y} \min_{x\in \mathcal X} \Phi(x,y)$ and  a solution $x^*$ of $ \min_{x_\in \mathcal X} \Phi(x,y^*)$, that is, 
$x^*=\Pi_{\mathcal X}(x^*-\Phi'_x(x^*,y^*))$,  
such that the equation $y^*=\Pi_{\mathcal Y}(y^*+\Phi'_y(x^*,y^*))$  also holds, then $(x^*,y^*)$ is a saddle point of $\Phi$, which then leads to \black{the fact that} $x^*$ is a solution of the primal problem $\min_{x\in \mathcal X}\max_{y\in \mathcal Y} \Phi(x,y)$.    
This motivates us to use a dual subgradient method that solves the dual problem of \eqref{primal}, which is the maximization problem of a concave function. \black{This is described in the next subsection.} 

\subsection{A dual subgradient method}

Define the functions $L(W,H;\lambda)=\bar{D}_{\Omega}^\lambda(X,WH)$, and
$g(\lambda) = \min_{(W,H) \geq 0} L(W,H;\lambda)$, and  the set 
\[
\Lambda=\{\lambda: \lambda\geq 0, \|\lambda\|_1 = 1\}. 
\] 
It then follows from Danskin's theorem \cite[Proposition B.25]{Bertsekas2009} that the vector  $-g'(\lambda)$ with 
\begin{equation}
\label{subgrad}
g'(\lambda)=\left[\bar D_\beta (X,W_\lambda H_\lambda) \right]_{\beta \in \Omega},  
\end{equation} 
 where $(W_\lambda, H_\lambda)\in\argmin_{W\geq 0,H\geq 0} \bar D_\Omega^\lambda(X,W H)$, is a subgradient of $-g$ at $\lambda$.
\hien{We now solve the dual problem}
\begin{equation}
\label{dual}
\max_{\lambda\geq 0, \|\lambda\|_1 = 1} g(\lambda), 
\end{equation}  
to obtain an optimal solution $\lambda^*$. 
We observe that $g(\lambda)$ is  concave and, as such, a subgradient method with \black{a suitable choice of step sizes} guarantees the convergence to a global optimal solution of the concave maximization problem in~\eqref{dual}; 
see, for example,~\cite{anstreicher2009two}.  

Algorithm~\ref{algo:subgrad} describes a dual subgradient method. It is worth noting that Problem~\eqref{dual} can also be solved by a mirror descent method; see for example \cite[Chapter 4]{Bubeck2015}. 
\algsetup{indent=2em}
\begin{algorithm}[ht!]
\caption{A dual subgradient method} 
\label{algo:subgrad}
\begin{algorithmic}[1]
\REQUIRE
The matrices $X \in \mathbb{R}^{m \times n}_+$, 
a finite set $\Omega$ of nonnegative real numbers. 

\ENSURE $(W,H)$, an approximate solution to~DR-NMF~\eqref{drnmf} . 
\medskip 

\STATE Initialize $\lambda^{(1)}_\beta = \frac{1}{|\Omega|}$ for all $\beta \in \Omega$. 

\FOR{$k=1,2,\dots$}
\STATE \label{algo2:setp3} Compute $(W^{(k)},H^{(k)})$ as an optimal solution to 
\begin{equation}
\label{eq:primal}
 \min_{(W,H) \geq 0} \bar D_\Omega^{\lambda^{(k)}}(X,W H). 
\end{equation}  

\STATE \label{algo2:setp4} Update \[
\lambda^{(k+1)}={\Pi}_{\Lambda}\big(\lambda^{(k)}+\rho_k g'(\lambda^{(k)})\big), 
\] 
 where $g'(\lambda^{(k)})$ is computed as in  \eqref{subgrad}.
\ENDFOR  

\end{algorithmic}
\end{algorithm} 
Note that we take $\Pi_\Lambda$ in Step 4 of Algorithm~\ref{algo:subgrad} to be the Euclidean projection operator. This allows us to establish a convergence guarantee for  the sequence of dual parameters $\{\lambda^{(k)}\}_{k\in\mathbb{N}}$ generated in Step~\ref{algo2:setp4} of  Algorithm~\ref{algo:subgrad} by applying \cite[Theorem 3]{anstreicher2009two}.  \revise{Specifically, suppose the step sizes satisfy  $\rho_k\to 0$,  $\sum_{k=1}^\infty\rho_k = +\infty$ and $\sum_{k=1}^\infty \rho_k^2 <\infty$. 
Then the sequence $\{\lambda^{(k)}\}_{k\in\mathbb{N}}$  generated by Algorithm~\ref{algo:subgrad}  converges to a solution $\lambda^*$ of~\eqref{dual}. Furthermore, suppose $(W^*,H^*)$ is a limit point of $(W^{(k)},H^{(k)})$ and assume that $(W,H,\lambda) \mapsto L(W,H;\lambda)$ is lower semicontinuous at $(W^*,H^*,\lambda^*)$. Then we have  $(W^*,H^*)\in \arg\min_{(W,H)\geq 0} \bar{D}_{\Omega}^{\lambda^*}(X,WH)$. Indeed, let  $\{k_n\}_{n\in\mathbb{N}}$ be such that $(W^{(k_n)}, H^{(k_n)})\to (W^*,H^*)$ as $n\to\infty$. Step \ref{algo2:setp3} of Algorithm~\ref{algo:subgrad} implies that 
\[
L(W^{(k_n)},H^{(k_n)}; \lambda^{(k_n)}) 
\leq 
L(W,H; \lambda^{(k_n)}), 
\;\;\forall\,  (W,H) \geq 0.
\]
Taking $n\to\infty$, we obtain 
\begin{align*}
L(W^*,H^*; \lambda^*) & \leq \liminf_{n\to \infty} L(W^{(k_n)},H^{(k_n)}; \lambda^{(k_n)}) \\ 
& \leq L(W,H;\lambda^*) \text{ for all } (W,H) \geq 0. 
\end{align*}
 Hence $  (W^*,H^*)\in\arg\min_{(W,H)\geq 0} \bar{D}_{\Omega}^{\lambda^*}(X,WH)$. 
}



 \revise{We have shown that the iterates generated by Algorithm~\ref{algo:subgrad} converge} to a solution $\lambda^*$ of the dual problem and to a solution  $(W^*,H^*)$ of $\min_{(W,H)\geq 0} L(W,H; \lambda^*)$. After obtaining  $\lambda^*$, the discussion after Proposition~\ref{prop:SPP} indicates that if we assume that $L$ has a saddle point (which is a common assumption as mentioned in Section \ref{sec:related}) and that we can find a solution $(W^*,H^*)$ of $\min_{(W,H)\geq 0} L(W,H; \lambda^*)$ that satisfies the condition $\lambda^*=\Pi_{\Lambda}\big(\lambda^*+L'_{\lambda}(W^*,H^*; \lambda^*)\big),$ then we can recover a solution $(W^*,H^*)$ of the primal problem \eqref{primal}. 
Hence we can \black{regard the output $(W^{(k)},H^{(k)})$ of Algorithm~\ref{algo:subgrad} as} an approximate solution of \eqref{primal}. In practice, we can run Algorithm~\ref{algo:subgrad} until we observe that the change between two consecutive \black{iterates} is \black{negligible}; for example stop the algorithm when 
$
\|\lambda^{(k+1)}-\lambda^{(k)}\|_1 \; \leq \;  \varepsilon, 
$
for a predetermined \black{tolerance} $\varepsilon>0$. Since $\|\lambda^{(k)}\|_1 = 1$ for all $k$, choosing for example $\varepsilon = 0.001$ means that we stop the algorithm when $\lambda^{(k)}$ is   modified by less than 0.1\% (compared to the previous iterate).

 The performance of Algorithm~\ref{algo:subgrad} critically depends on the solver for solving the weighted-sum minimization in Step~\ref{algo2:setp3}, which itself is a difficult non-convex optimization problem. We can use Algorithm~\ref{wsnmf} to  find  \rd{an approximate solution} $(W^{(k)},H^{(k)})$ 
 in Step~\ref{algo2:setp3}. However, subgradient methods are often slow in practice. \black{Indeed, we observe that  Algorithm~\ref{algo:subgrad} combined with Algorithm~\ref{wsnmf} is very slow for the data sets we use in our experiments (see Section~\ref{sec:compare_sg_heur} and  Figure~\ref{fig:alg2vssubgr}).} 
 Therefore, although Algorithm~\ref{algo:subgrad} provides some convergence \black{guarantees} for \black{the} primal and dual problems, we are \black{motivated} to propose another practical approach for \rd{finding an  approximate solution to} \eqref{drnmf}. In the following, we present a heuristic scheme that performs very well, significantly better than Algorithm~\ref{algo:subgrad} where Step~\ref{algo2:setp3} is approximately solved via Algorithm~\ref{wsnmf}.

 \color{black} 
 

\subsection{A  \color{black} Frank-Wolfe \color{black} heuristic scheme for DR-NMF}

\color{black}
 
Unfortunately, Algorithm~\ref{algo:subgrad} is not practical because Step~\ref{algo2:setp3} requires one to solve~\eqref{eq:primal} which is NP-hard in general (since it is a generalization of NMF).  
Hence we instead propose a heuristic scheme described in Algorithm~\ref{algo:drnmf}. 

\algsetup{indent=2em}
\begin{algorithm}[ht!]
\caption{Heuristic \color{black} Frank-Wolfe \color{black} scheme for DR-NMF~\eqref{drnmf}} \label{algo:drnmf}
\begin{algorithmic}[1]
\REQUIRE
The matrices $X \in \mathbb{R}^{m \times n}_+$ 
and 
an initialization $W^{(0)} \in \mathbb{R}^{m \times r}_+$ 
and $H^{(0)} \in \mathbb{R}^{m \times r}_+$, 
a finite set $\Omega$ of nonnegative real numbers.   

\ENSURE $(W,H)$, an approximate solution to~DR-NMF~\eqref{drnmf}.
\medskip 

\STATE Initialize $\lambda_\beta = \frac{1}{|\Omega|}$ for all $\beta \in \Omega$. 

\FOR{$k=1,2,\dots$}

\STATE \label{algo3:setp3}  Update $H^{(k+1)}$ using Algorithm~\ref{muHwsnmf} with initialization $H = H^{(k)}$, and with $W = W^{(k)}$ and $\lambda = \lambda^{(k)}$.

\STATE \label{algo3:setp4}  Update $W^{(k+1)}$ using Algorithm~\ref{muHwsnmf} and the symmetry of the NMF problem, with initialization $W = W^{(k)}$, and with $H = H^{(k+1)}$ and $\lambda = \lambda^{(k)}$. 

\STATE \label{algo3:setp5} Let 
$\beta^* \in \argmax_{\beta \in \Omega} \bar{D}_{\beta}(X,W^{(k+1)}H^{(k+1)})$ and 
\begin{equation}
(\lambda^{(k)}_*)_\beta = \left\{ \begin{array}{cc}
 1 & \text{ if } \beta = \beta^*, \\ 
 0 & \text{ if } \beta \neq \beta^*. 
\end{array} \right.  \label{eqn:lambda_star}
\end{equation}

\STATE 
Choose a step size $\gamma_k$, and update 
\[
\lambda^{(k+1)} 
 = \lambda^{(k)}  +  \gamma_k (\lambda_*^{(k)} -\lambda^{(k)}). 
\] 



\ENDFOR  

\STATE $(W,H) = ( W^{(k+1)} , H^{(k+1)}  )$. 

\end{algorithmic}
\end{algorithm}

Let us explain the main ideas behind Algorithm~\ref{algo:drnmf}. Steps~\ref{algo3:setp3} and~\ref{algo3:setp4} are {designed to} decrease $(W,H)\mapsto L(W,H;\lambda^{(k)})$.  Note that if we perform Steps~\ref{algo3:setp3} and~\ref{algo3:setp4}  {repeatedly}, its output will approximate the output of Step~\ref{algo2:setp3} of Algorithm~\ref{algo:subgrad}.  However, this would be computationally rather expensive as it would require many updates of $(W,H)$ for each $\lambda^{(k)}$, 
and the MU constitutes   the most expensive steps of Algorithm~\ref{algo:drnmf}. 

\color{black} 
The descent direction $\lambda_*^{(k)}$ used to update $\lambda$ at iteration $k$ is the one from the Frank-Wolfe (FW)  algorithm~\cite{FW56}, \color{black}  
 which is  also known as the conditional gradient method;  see for example~\cite{Jaggi2013} and the references therein. 
  In the context of solving DR-NMF~\eqref{drnmf}, let us explain the intuition behind the descent direction $\lambda^{(k)}_*$ , defined in Step~\ref{algo3:setp5} of Algorithm~\ref{algo:drnmf}.   
 Letting  
$\beta^* \in \argmax_{\beta \in \Omega} \bar{D}_{\beta}(X,W^{(k+1)}H^{(k+1)})$, we have  for all $\beta\in\Omega$ that 
$$
\bar{D}_{\beta^*}(X,W^{(k+1)}H^{(k+1)}) \geq \bar{D}_{\beta}(X,W^{(k+1)}H^{(k+1)}). 
$$ 
Defining $\lambda^{(k)}_*$ as the vector with a single non-zero entry equal to one at position $\beta^*$, see~\eqref{eqn:lambda_star}, we have 
$
\lambda^{(k)}_* = \argmax_{\lambda \in \Lambda} \bar{D}_{\beta}^\lambda(X,W^{(k+1)}H^{(k+1)}).
$
 Therefore, since we are trying to solve  the optimization problem
$\min_{(W,H) \geq 0} \max_{\beta \in \Omega} \bar{D}_{\beta}(X,WH)$, 
the $\beta^*$-divergence should be given more importance at the next iteration {in order} for $\bar{D}_{\beta^*}(X,WH)$ to decrease, 
and hence the maximum among the $\beta$-divergences to decrease as well at the next iteration. 
Finally, Algorithm~\ref{algo:drnmf} updates $\lambda^{(k)}$ in Step~6 as follows   
\begin{align}
\lambda^{(k+1)} 
& = \lambda^{(k)}  +  \gamma_k (\lambda_*^{(k)} -\lambda^{(k)}),  \label{eqn:fw} \\ 
& = (1-\gamma_k) \lambda^{(k)}  +   \gamma_k \lambda_*^{(k)}. \nonumber 
\end{align} 
\color{black}
In our experiments, we choose the step sizes 
$\gamma_k = \frac{1}{k+1}$.  
 We leave the fine-tuning of the step sizes as a future direction of research, although we have tried different step sizes, 
 and we were not able to find step sizes that perform  significantly better than $\gamma_k = \frac{1}{k+1}$. 
 In fact, the performance for $\gamma_k$ of similar order is very similar. For example, the standard FW parameter choice of 
 $\gamma_k = \frac{2}{k+2}$ which yields essentially the same but slightly worse performance. 
 
\color{black}


\textcolor{black}{We emphasize that Algorithm~\ref{algo:drnmf} is a  {heuristic algorithm}, and we leave {convergence guarantees} as a research topic for future work. 
We use Algorithm~\ref{algo:drnmf} in our experiments and observe that it performs very well in {terms} of {simultaneously  decreasing all $\beta$-divergences for $\beta\in\Omega$. Thus, we do not need prior knowledge on the noise distribution or, equivalently, on the value of $\beta$.}
}
 \color{black}

\subsection{Comparison between the dual subgradient method and the heuristic scheme for DR-NMF}  \label{sec:compare_sg_heur}

The update in Step~\ref{algo3:setp5} of Algorithm~\ref{algo:drnmf} {\color{black} results in} faster convergence compared to the subgradient direction (see  Step~\ref{algo2:setp4} of  Algorithm~\ref{algo:subgrad}) 
because it gives  much more importance to the $\beta$-divergence that is  maximal at the current iteration.  
In fact, the entries of the subgradient are always all positive 
(unless $X=WH$ in which case the problem is solved). 
 Thus, the direction that places weight {\em only} at the current maximum $\beta$-divergence (as is done in Step~\ref{algo3:setp5} of Algorithm~\ref{algo:drnmf}) outperforms the subgradient direction empirically, leading to much faster convergence in practical problems.

Let us compare the dual subgradient method and the heuristic scheme for DR-NMF on a simple synthetic data set. However, we have made the same observations on all other data sets we have experimented with such as the ones presented in Section~\ref{sec:ne}. 
\begin{figure}[t]
\begin{center}
\includegraphics[width=0.48\textwidth]{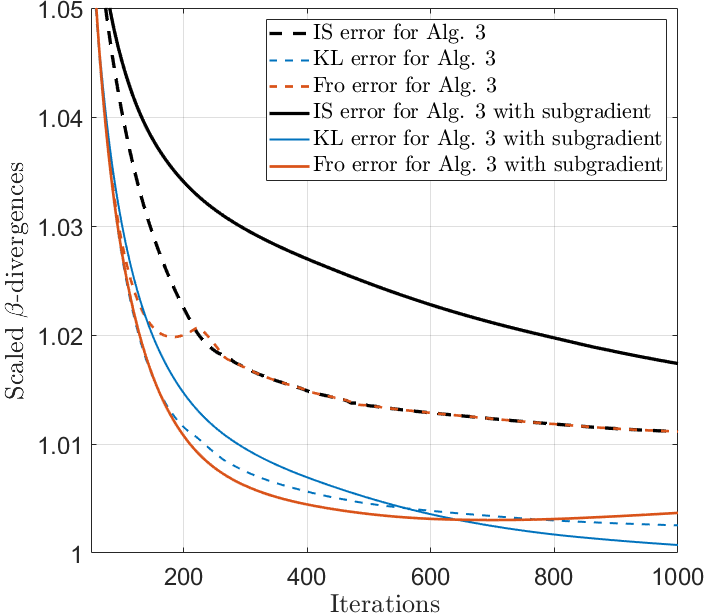}   
\caption{Comparison of Algorithm~\ref{algo:drnmf} with its variant where Step \ref{algo3:setp5} is replaced by a standard subgradient step (i.e., Step \ref{algo2:setp4} of Algorithm~\ref{algo:subgrad}). The curves display the evolution of the scaled $\beta$-divergences (referred to as ``error'' in the legend).  \label{fig:alg2vssubgr}}
\end{center}
\end{figure} 
Figure~\ref{fig:alg2vssubgr} illustrates the distinction between the two algorithms  with a synthetic  experiment  that compares   Algorithm~\ref{algo:drnmf} with the variant in which Step $5$ is replaced with a standard subgradient step, that is,  Step~\ref{algo2:setp4} of Algorithm~\ref{algo:subgrad}. 
In this illustrative experiment, the entries of a $100$-by-$100$ matrix $X$ are generated uniformly at random in the interval $[0,1]$, and we use $r=10$ and $\Omega = \{0,1,2\}$, that is,  DR-NMF with the IS-divergence, the KL-divergence and the Frobenius norm. Both variants are initialized with the same matrices  
$(W^{(0)}, H^{(0)})$ whose entries are also generated uniformly at random in $[0,1]$. 
 
We observe that 
the variant using the subgradient converges very slowly. Indeed, the maximum of the three objectives (the IS-divergence) is far from convergence, even after $1000$ iterations.\footnote{It required $3000$ iterations to make the IS-divergence and the Frobenius norm intersect, but then the Frobenius norm becomes larger and, within the next  $7000$ iterations (for a total of $10000$ iterations), the IS divergence remains larger hence convergence is not attained.}
In contrast,  Algorithm~\ref{algo:drnmf} converges much faster. In particular, the values of the IS-divergence and the Frobenius norm quickly converge to one another. 
All in all, Algorithm~\ref{algo:drnmf} finds a solution with scaled $\beta$-divergence within 2\% of the smallest possible values for the three $\beta$-divergences within $240$ iterations, that is,  $\max_{\beta\in\Omega}\bar{D}_\beta(X,W^{(k)}H^{(k)}) \leq 1.02$ for $k \geq 240$. 
This is made possible because of our more aggressive heuristic strategy to update $\lambda$. In contrast, if one uses the subgradient direction, about~$800$ iterations are required to obtain the same approximation guarantee.

\begin{remark}[Property of DR-NMF solutions] 
We observe on Figure~\ref{fig:alg2vssubgr} that the two scaled $\beta$-divergences with the largest values are equal to each another. The same behaviour will often be observed in the extensive sets of experiments in Section~\ref{sec:ne}. Let us explain why this is expected to happen. First, recall that the lowest possible value of a single scaled $\beta$-divergence is one; see Section~\ref{sec:scaling}. Second, the maximum scaled $\beta$-divergence is attained at a single scaled $\beta$-divergence when it is strictly larger than all the other scaled $\beta$-divergences. In that case, the maximum scaled $\beta$-divergence  will be larger than one, and hence it can typically\footnote{Because of the non-convexity of the objectives of DR-NMF, such a descent direction is not guaranteed to exist.} be reduced locally while ensuring that the other $\beta$-divergences remain smaller, hence reducing the maximum scaled $\beta$-divergence.  
\end{remark}

\section{Numerical Experiments}  \label{sec:ne} 

In this section, we apply 
DR-NMF on several data sets. In all cases, we perform 1000 iterations. All tests are preformed using Matlab
R2015a on a laptop Intel CORE i7-7500U CPU @2.9GHz 24GB RAM. 
The code is available on Code Ocean via \url{https://doi.org/10.24433/CO.7769595.v1}.

\subsection{MO-NMF: Examples of the Pareto frontier on synthetic data} \label{sec:pareto}

In this section, we illustrate the use of Algorithm~\ref{muHwsnmf} to compute Pareto-optimal solutions. We will focus on the case $\beta=0,1,2$, that is, IS- and KL-divergences and the Frobenius norm. Note, however, that our algorithm and code can deal with any $\beta \geq 0$ and any finite set $\Omega$. 

We generate the input matrix $X$ as follows: $X = \max \big( 0 ,  \tilde{W}\tilde{H} + N \big)$ where the component matrices $\tilde{W}$, $\tilde{H}$ and the noise matrix $N$ are generated as follows:  
\begin{itemize}
\item The entries of $\tilde{W} \in \mathbb{R}^{200 \times 10}$ and 
$\tilde{H} \in \mathbb{R}^{10 \times 200}$ are generated using the uniform distribution in the interval [0,1]. We define $\tilde{X} = \tilde{W}\tilde{H}$ which is the noiseless low-rank matrix. 

\item Let us define $x_\beta = 1$ if $\beta \in \Omega$, $x_\beta=0$ otherwise. Let also 
\[
\tilde{N} = x_0 
\frac{N_{\text{IS}}}{\|N_{\text{IS}}\|_F} 
+ x_1 \frac{N_{\text{KL}}}{\|N_{\text{KL}}\|_F} 
+ x_2 \frac{N_{\text{F}}}{\|N_{\text{F}}\|_F} , 
\] 
where 
\begin{itemize}

\item $N_{\text{IS}} = \tilde{X} \circ G$ is multiplicative Gamma noise where each entry of $G$ is generated using the normal distribution of mean $0$ and variance $1$, 

\item each entry of $N_{\text{KL}}$ is generated according to the Poisson distribution of parameter $1$ \ngi{(for simplicity, since the expected value of $(\tilde{W}\tilde{H})_{i,j}$ is the same for all $(i,j)$)},

\item each entry of $N_{\text{F}}$ is generated using the normal distribution of mean $0$ and variance $1$. 

\end{itemize}
We set $N = \epsilon  \frac{\|\tilde{X}\|_F}{\|\tilde{N}\|_F} \tilde{N}$ with $\epsilon = 0.2$. 
\end{itemize}
Finally, $X = \max(0, \tilde{X}+ N)$ is a low-rank matrix to which had been contaminated with  20\% of noise (that is, $\|N\|_F = 0.2 \|\tilde{X}\|_F$) and then was projected onto the nonnegative orthant. The noise is constructed using the distributions corresponding to $\beta \in \Omega$.

Figure~\ref{fig:synt} shows the Pareto-optimal solutions for MO-NMF. More precisely, it provides the solution for the problems
\[
\min_{(W,H) \geq 0} \bar{D}_{\Omega}^\lambda(X,WH), 
\]
where $\lambda = (\ell,1-\ell)$ for $\ell=0,0.1,\dots,1$, and for $\Omega = \{0,1\},  \{0,2\},  \{1,2\}$. To simplify computation, we have used the true underlying solution $(W_{\mathrm{t}},H_{\mathrm{t}})$ as the initialization (using random or other initializations sometimes generate solution which are more often not on the Pareto frontier because NMF may have many local minima).  
The Pareto frontier is as expected: the smallest possible value for each objective is 1 (because of the scaling), for which the other objective function is the largest. As $\lambda$ changes, one objective increases while the other decreases. The DR-NMF solution computed with 
Algorithm~\ref{algo:drnmf} finds the point on the Pareto frontier  such that 
$\bar{D}_{\beta_2}(X,WH) = \bar{D}_{\beta_1}(X,WH)$ for $\beta_1\neq \beta_2 \in \Omega$. 
\begin{figure}[ht!]
\begin{center}
\begin{tabular}{c}
\includegraphics[width=0.35\textwidth]{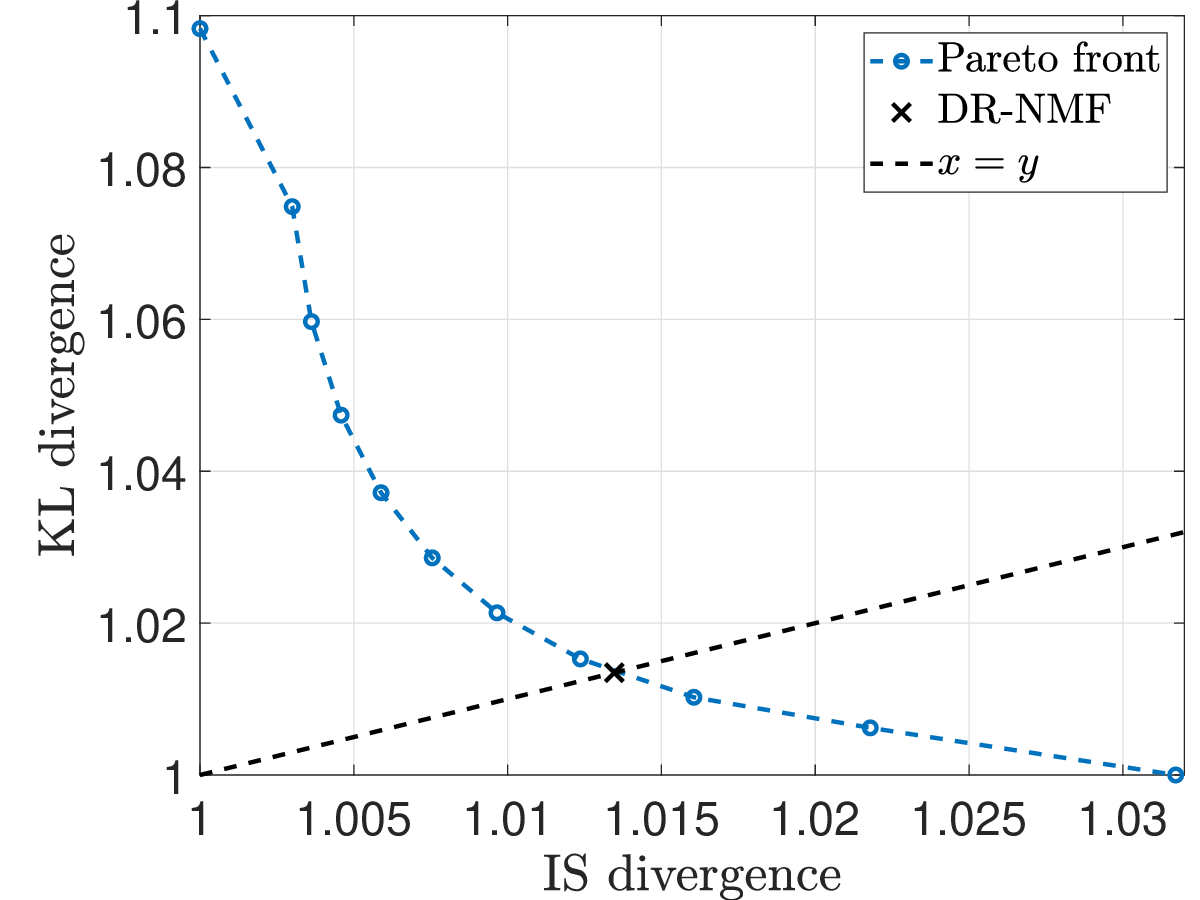}  \\ 
\includegraphics[width=0.35\textwidth]{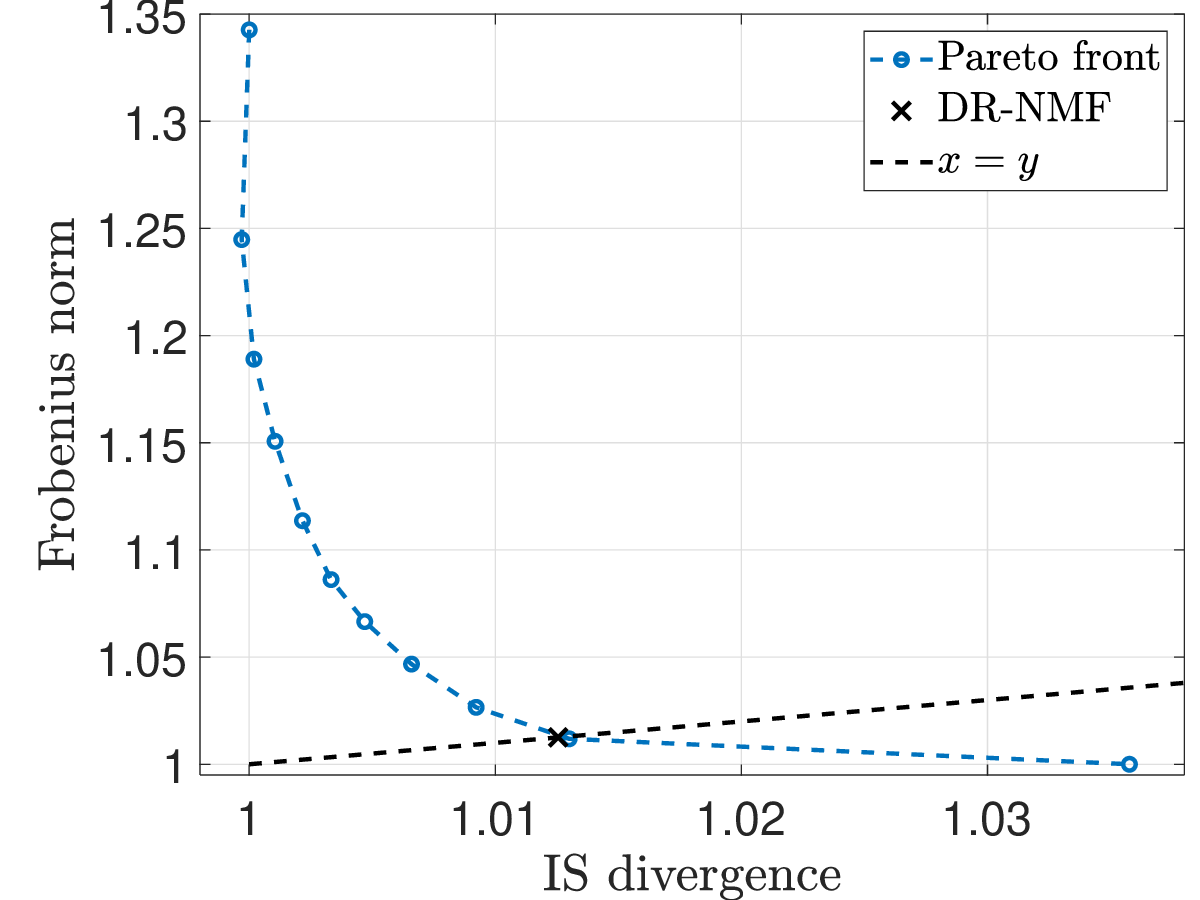} \\ 
\includegraphics[width=0.35\textwidth]{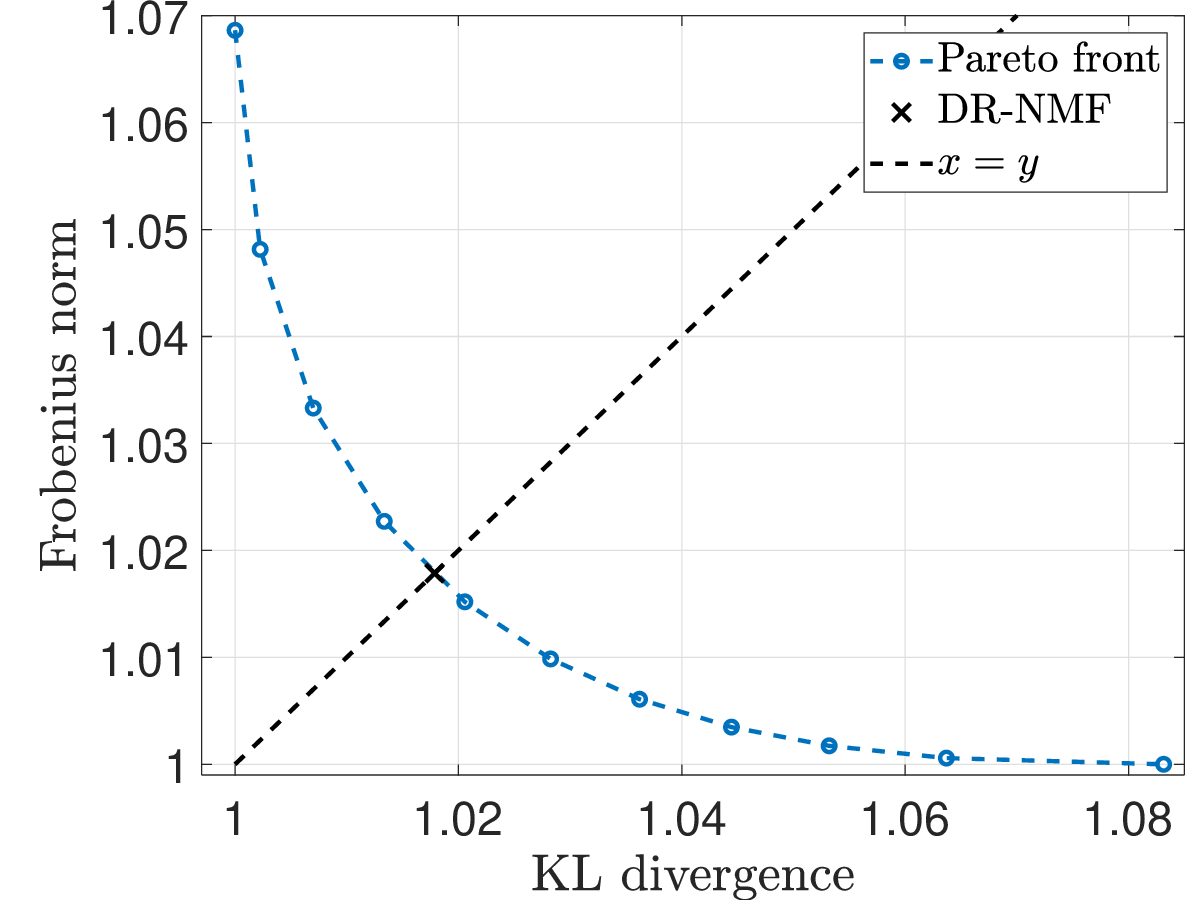}  
\end{tabular}
\caption{ Pareto-optimal solutions for $\lambda = (\ell,1-\ell)$ for $\ell=0,0.1,\dots,1$ and $\Omega = \{0,1\},  \{0,2\},  \{1,2\}$ (top to bottom), 
and solution computed by DR-NMF (Algorithm~\ref{algo:drnmf}).  \label{fig:synt}}
\end{center}
\end{figure} 

For DR-NMF, we observe that 
\begin{itemize}

\item The solution of DR-NMF does not necessarily coincide with a value of $\lambda$ close to $(0.5,0.5)$. For example, for the case of the IS divergence with the Frobenius norm, it is close to $\lambda = (0.9,0.1)$. 

\item Using DR-NMF allows to obtain a solution with low error for both objectives, always at most 2\% worse than the lowest error. Minimizing a single objective sometimes leads to solution with error up to 35\% higher than the lowest (in the case IS divergence with Frobenius norm). We will observe a similar behaviour on real data sets. 

\end{itemize}

\subsection{Sparse document data sets: $\Omega = \{1,2\}$}  \label{sec:sparse}

For sparse data sets, 
it is known that only the $\beta$-divergence for $\beta=1,2$ can exploit the sparsity structure. 
In fact, in all other cases, all entries of the product $WH$ have to be computed explicitly which is impractical for large sparse matrices since $WH$ can be dense. 
In other words, let $K$ denote the number of non-zero entries of $X$. Then the MU for NMF with the $\beta$-divergence for $\beta = 1,2$ can be run in $O(Kr)$ operations, while for the other values of $\beta$, it requires $O(mnr)$ operations. 

As explained in~\cite{chi2012tensors}, 
for \ngi{sparse word-count} matrices, 
Poisson noise is the most appropriate model; in fact, Gaussian noise (and any dense noise) does not make much sense on sparse data sets. Hence we expect KL-NMF to provide better results than Fro-NMF. 
\ngi{However, we believe it is rather interesting to run DR-NMF with $\Omega = \{1,2\}$ on such data sets to see how it performs. One should expect DR-NMF to perform on average worse than KL-NMF (since it has to take into account the Frobenius norm which is not appropriate) but better than the Frobenius norm (since it takes into account the more appropriate KL-NMF). }

In this section, we use the 15 sparse document data sets from~\cite{ZG05}. These are large and highly sparse matrices whose entries $X(i,j)$ is the number of times word $j$ appears in document $i$. We apply KL-NMF, Fro-NMF and DR-NMF with $\Omega = \{1,2\}$. To simplify the comparison, reduce the computational load and to have a good initial solution, 
\color{black}
we use the same initial matrices $(W^{(0)},H^{(0)})$ in all cases, namely the solution obtained by the successive projection algorithm~\cite{Araujo01} that has provable guarantee under the separability condition~\cite{gillis2013fast, arora2013practical}.  
\color{black} 
We perform rank-$r$ factorization where $r$ is the number of classes reported for these data sets. 


\begin{center}
\begin{table*}[h!] 
\begin{center}
\caption{\color{black}Comparison of NMF with KL-divergence and Frobenius norm, and DR-NMF with $\Omega = \{1,2\}$ on text mining data sets from~\cite{ZG05}. 
Bold numbers indicate the best accuracy, underlined numbers indicate the second best accuracy. \label{tab:dtm}}   
 \begin{tabular}{|c|c|ccc|cc|cc|}  \hline 
Data set & $r$ & \multicolumn{3}{c}{Clustering accuracy (\%)}  
& \multicolumn{2}{|c|}{$\bar{D}_{1}(X,WH) - 1$ (\%)}
&  \multicolumn{2}{|c|}{$\bar{D}_{2}(X,WH) - 1$ (\%)} \\ 
&   & KL-NMF & Fro-NMF & DR-NMF  & Fro-NMF & DR-NMF & KL-NMF  & DR-NMF 
        \\ \hline 
NG20 &  20 &  \textbf{42.15}  &  23.08  &  \underline{28.74}   
 &  21.47 &  3.85 
 &  149.48 &  3.83 
 \\ \hline 
ng3sim &   3 &  \textbf{63.48}  &  38.06  &  \underline{49.87}  
&  16.82 &  2.70 
&  17.32 &  2.70 
\\ \hline 
classic &   4 &  \textbf{83.66}  &  55.64  &  \underline{78.46}  
&  13.19 &  0.74 
&  2.44 &  0.74 
\\ \hline 
ohscal &  10 &  \textbf{37.45}  &  30.50  
&  \underline{32.13}  
   &  10.03 &  1.76 
   &  9.60 &  1.75 
   \\ \hline 
k1b &   6 &  \textbf{64.27}  &  59.19  &  \underline{60.30}  
   &  9.00 &  1.32 
   &  5.02 &  1.32 
   \\ \hline 
hitech &   6 &  43.29  &  \underline{46.94}  &  \textbf{48.02} 
 &  8.27 &  1.12 
 &  3.98 &  1.13 
 \\ \hline 
reviews &   5 &  \textbf{75.65}  &  51.19  &  \underline{74.88}  
&  7.89 &  1.03 
 &  7.70 &  1.03 
 \\ \hline 
sports &   7 &   \underline{43.93}  &  40.37  &  \textbf{50.26}   
 &  9.60 &  1.24 
 &  7.10 &  1.24 
 \\ \hline 
la1 &   6 &  \underline{65.95}  &  65.04  &  \textbf{67.98} 
 &  9.26 &  1.03 
 &  3.61 &  1.03 
 \\ \hline  
la12 &   6 &  \textbf{56.25}  &  54.80  &  \underline{54.29}
 &  7.32 &  0.70 
 &  2.76 &  0.70 
 \\ \hline 
la2 &   6 &  \textbf{54.96}  &  49.17  &  \underline{52.07}  
 &  9.21 &  0.82 
 &  3.04 &  0.82 
 \\ \hline 
tr11 &   9 &  \textbf{62.32}  &  50.48  &  \underline{51.45} 
  &  22.88 &  4.48 
  &  97.27 &  4.47 
  \\ \hline 
tr23 &   6 &  34.80  &  \underline{35.29}  &  \textbf{38.73}  
 &  56.04 &  3.83 
  &  47.36 &  3.78 
  \\ \hline 
tr41 &  10 &  \textbf{54.33}  &  44.99  &  \underline{53.08} 
 &  24.38 &  4.93 
 &  46.17 &  4.90 
 \\ \hline 
tr45 &  10 &  \textbf{46.81}  &  38.26  &  \underline{39.13} 
  &  42.52 &  10.15 
  &  50.14 &  10.15 
  \\ \hline 
\hline 
Average &  &   55.29   &  45.53  &  51.96 
 & 
 17.86 &  2.65 
 &  30.20 &  2.64 
 \\ \hline 
\end{tabular}  
\end{center}
\end{table*}
\end{center}  

Table~\ref{tab:dtm} reports the results. The first and second columns  report the name of the data set and the number of classes, respectively. 
The next \color{black} four \color{black} columns report the accuracies of the clustering obtained with the factorizations $(W,H)$ produced by 
KL-NMF, Fro-NMF, and DR-NMF with $\Omega = \{1,2\}$ \color{black} solved via  Algorithm~\ref{algo:drnmf}.  \color{black} 
Given the true disjoint clusters 
$C_i \subset \{1,2,\dots,m\}$ for $1 \leq i \leq r$ and given a computed disjoint clustering $\{\tilde{C}_i\}_{i=1}^r$, 
its {\em accuracy} is defined as 
\[
\text{accuracy}\left( \{\tilde{C}_i\}_{i=1}^r \right)
\; = \;  
\min_{\pi \in [1,2,\dots,r]} \frac{1}{m} \sum_{i=1}^r |C_i \cap \tilde{C}_{\pi(i)}|,  
\]
where $[1,2,\dots,r]$ is the set of permutations of $\{1,2,\dots,r\}$. 
For simplicity, given an NMF $(W,H)$ where each row of $H$ corresponds to a topic, 
\color{black} 
we cluster the documents by selecting its closest topic, that is, document $j$ is assigned to the topic $k$ that maximizes $\frac{X(i,:)^T H(k,:)}{\| H(k,:)\|_2}$.  
\color{black} 
The next \color{black} three \color{black} columns report how much higher the KL error (in percent) of the solutions 
of Fro-NMF and DR-NMF  
are compared to KL-NMF, that is, it reports 
\[
\bar{D}_{1}(X,WH) - 1 = 
\frac{ \mathrm{KL}(X,WH)  }{ \mathrm{KL}(X,W_1H_1) } - 1 
\]
where $(W_1,H_1)$ is the solution computed by KL-NMF. 
The last \color{black} three \color{black} columns report how much higher the Frobenius error (in percent) of the solutions of KL-NMF and DR-NMF are compared to Fro-NMF, that is, 
\[
\bar{D}_{2}(X,WH) - 1 = 
\frac{ \|X-WH\|_F^2  }{ \|X-W_2H_2\|_F^2 } - 1 
\]
where $(W_2,H_2)$ is the solution computed by Fro-NMF.

We observe the following: 
\begin{itemize}

\color{black}


\item In terms of clustering, 
DR-NMF in fact allows us to be robust in the sense that it is able to provide \emph{in all cases} at least the second highest clustering accuracy. 
On four data sets, it is even able to provide the highest accuracy, sometimes by a large margin.  
Globally, DR-NMF does not perform as well as KL-NMF although on average their accuracy only differs by 3.32\%. However, DR-NMF performs better than Fro-NMF, with 6.44\% higher accuracy on average.

\item In terms of error, as already noted in the previous section, DR-NMF is able to simultaneously provide solutions with small KL and Frobenius error, on average 2.65\% higher than the solution computed with a single objective. On the other hand, optimizing a single objective often leads to very large errors for the other one, 
up to 149\% on NG20, with an average of 17.86\% for Fro-NMF and 30.20\% for KL-NMF.  

\end{itemize}

\subsection{Dense time-frequency matrices of audio signals: $\Omega = \{0,1\}$}   \label{sec:audio}

NMF has been used successfully to separate sources from a single audio recording. 
However, there is a debate in the literature as to whether the KL or the IS divergence should be used; see~\cite{virtanen2007monaural, fevotte2009nonnegative} and the references therein. 
In fact, as we will see, IS-NMF and KL-NMF provide rather different results on different audio data sets. 
On one hand, due to its insensitivity to scaling (see Section~\ref{sec:scaling}), IS-NMF gives the same relative importance to all entries of the data matrix. For example, the error for approximating 1 by 10 is the same as for approximating 10 by 100, that is, $D_{0}(1,10)=D_{0}(10,100)$.  
On the other hand, KL-NMF gives more importance to larger entries as it is (linearly) sensitive to scaling; for example, 
the error for approximating 1 by 10 is ten times smaller than approximating 10 by 100, that is, $10 D_{1}(1,10)=D_{1}(10,100)$.  

\subsubsection{Quantitative results} 

\begin{center}  
 \begin{table*}[h!] 
 \begin{center} 
\caption{\color{black} Comparison of NMF with the IS- and KL-divergences, and DR-NMF with $\Omega = \{0,1\}$ on audio data sets with $m=149$ and $r=10$.  
The table reports the averages and standard deviations over 10 initializations. \label{tab:audio}} 
 \begin{tabular}{|c|c|cc|cc|}  \hline 
Data set & $n$ & \multicolumn{2}{|c|}{$\bar{D}_{0}(X,WH) - 1$ (\%)} & \multicolumn{2}{|c|}{$\bar{D}_{1}(X,WH) - 1$ (\%)} \\ 
 &   & KL-NMF & 
 {DR-NMF}  
 & IS-NMF &  {DR-NMF} \\ 
 \hline 
syntBassDrum &  543  &  39.54 $\pm$  3.28 &  7.06 $\pm$  3.01 
&  108.09 $\pm$  16.78 &  7.06 $\pm$  3.01 
\\ \hline 
piano$\_$Mary &  586  &  387.51 $\pm$  253.67 &  9.73 $\pm$  2.74 
&  177.71 $\pm$  22.79 &  9.73 $\pm$  2.74 
\\ \hline 
prelude$\_$JSB &  2582  &  31.81 $\pm$  3.57 &  13.05 $\pm$  3.55 
&  185.93 $\pm$  54.84 &  13.04 $\pm$  3.55 
\\ \hline 
syntCCcyGC &  1377  &  9.79 $\pm$  0.64 &  2.63 $\pm$  0.38 
&  42.21 $\pm$  10.29 &  2.63 $\pm$  0.38 
\\ \hline 
trio$\_$Brahms &  14813  &  360.49 $\pm$  44.65 &  14.74 $\pm$  1.99 
&  257.61 $\pm$  130.29 &  14.74 $\pm$  1.99 
\\ \hline 
trio$\_$bapitru &  6200  &  354.66 $\pm$  25.31 &  9.16 $\pm$  2.03 
&  249.99 $\pm$  28.72 &  9.15 $\pm$  2.03 
\\ \hline 
voice$\_$cell &  2181  &  186.46 $\pm$  2.20 &  13.98 $\pm$  3.75 
&  191.12 $\pm$  23.43 &  13.98 $\pm$  3.75 
\\ \hline 
ShanHur$\_$sunrise &  4102  &  53.51 $\pm$  8.32 &  12.31 $\pm$  1.36 
 &  184.72 $\pm$  34.74 &  12.30 $\pm$  1.35 
\\ \hline 
sisec$\_$mixdrums &  1249  &  25.61 $\pm$  0.87 &  12.74 $\pm$  1.38 
&  292.40 $\pm$  56.62 &  12.74 $\pm$  1.38 
\\ \hline 
sisec$\_$mixfemale &  1249  &  37.68 $\pm$  2.88 &  12.12 $\pm$  0.85 
 &  100.67 $\pm$  8.88 &  12.12 $\pm$  0.85 
\\ \hline 
\hline 
Average &  &  148.71 $\pm$  34.54 &  10.75 $\pm$  2.10 
&  179.05 $\pm$  38.74
 &  10.75 $\pm$  2.10 
 \\ \hline 
\end{tabular} 
 \end{center} 
 \end{table*} 
 \end{center}

Our DR-NMF approach overcomes the issue of having to choose between the IS- and KL-divergences by generating solutions which possess small IS and KL errors simultaneously. 
We use  10 diverse audio data sets: 
\begin{itemize}
\item voice$\_$cell, syntBassDrum and syntCCcyGC were downloaded from \url{http://isse.sourceforge.net/demos.html}. 

\item prelude$\_$JSB is the the well-tempered Clavier performed by Glenn Gould 1/13 between the 19th et 49th seconds, 
downloaded from \url{https://www.youtube.com/watch?v=IrJjPYi_vhM}. 

\item ShanHur$\_$sunrise was downloaded from \url{http://bass-db.gforge.inria.fr/fasst/}.  

\item trio$\_$Brahms and trio$\_$bapitru were derived from the TRIOS data set~\cite{fritsch2012high}; see \url{https://c4dm.eecs.qmul.ac.uk/rdr/handle/123456789/27}. 

\item  sisec$\_$mixdrums and sisec$\_$mixfemale come from the SISEC data set; see \url{http://sisec.wiki.irisa.fr/tiki-indexbfd7.html?page=Underdetermined+speech+and+music+mixtures}. 

\item piano$\_$Mary is a recording at the third author's house. 

\end{itemize}

Table~\ref{tab:audio} reports the results, exactly as done in the last columns of Table~\ref{tab:dtm}, except that it also reports the standard deviation among 10 random initializations.

For these data sets, the results are even more striking than for the sparse text data sets in Section~\ref{sec:sparse}. In particular,
DR-NMF has on average an error higher by about 10\% compared to both IS-NMF and KL-NMF, while KL-NMF (resp.\@ IS-NMF) has on average an increase in IS error of 149\% (resp.\@ 179\%). \ngi{Moreover,  DR-NMF is more robust in the sense that its standard deviation is significantly lower. This shows that by taking into account different objectives, DR-NMF is less sensitive to initialization.} 

As we will see in the next section, using DR-NMF allows to obtain more robust results than using IS-NMF or KL-NMF alone.

\subsubsection{Qualitative results} \label{sec:qual}

In the previous section, we have shown quantitative results showing that DR-NMF is able to obtain solutions with low KL and IS divergence simultaneously. In this section, we investigate the data set piano$\_$Mary in more detail and show that DR-NMF also leads to better separation for three comparative studies described in detail below: 
(1) no noise added to the signal, 
(2) Poisson noise added and 
(3) Gamma noise added. 
This data set is the first 4.7 seconds of ``Mary had a little lamb''. The sequence is composed of three notes, namely, $E_{4}$, $D_{4}$ and $C_{4}$. The recorded signal is downsampled to $f_{s}=16000$Hz yielding $T=75200$ samples. The short-time Fourier transform (STFT) of the input signal $x$ is computed using a Hamming window of size $F=512$ leading to a temporal resolution of 32ms and a frequency resolution of 31.25Hz. We use 50\% overlap between
two frames, leading to $n=294$ frames and $m=257$ frequency bins. 
Figure \ref{fig:mary_representations} displays the musical score. 
\begin{figure}[h!]
\centering
\begin{subfigure}[b]{0.4\textwidth}
\includegraphics[width=\textwidth]{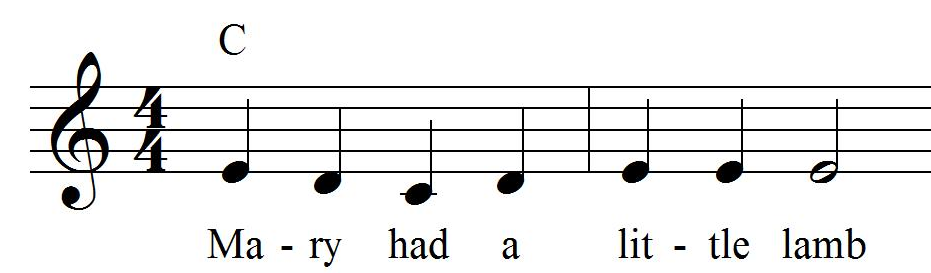}
    \end{subfigure}
\caption{Musical score of  ``Mary had a little lamb''. \ngi{The notes activate as follows: 
$E_4$, $D_4$, $C_4$, $D_4$, $E_4$, $E_4$, $E_4$.}
\label{fig:mary_representations}}
\end{figure}

There are three notes plus a fourth source. This last source is the very first offset of each note in the musical sequence, that is, some common mechanical vibration acting in the piano just before triggering a specific note, \reviselast{which can be associated to the hammer noise (denoted $H_N$)}, 
hence the correct rank is $r=4$; see~\cite{gillis2019learning} for more details.

\paragraph{No added noise}

Figure~\ref{fig:set1} displays the evolution of the \ngi{scaled} IS- and KL-divergences along iterations. 
DR-NMF is able to compute a solution with low IS and KL error, which is not the case of IS-NMF and DR-NMF (in particular, KL-NMF has IS error almost 9 times larger than IS-NMF).    
\begin{figure}
    \centering
        \includegraphics[width=0.5\textwidth]{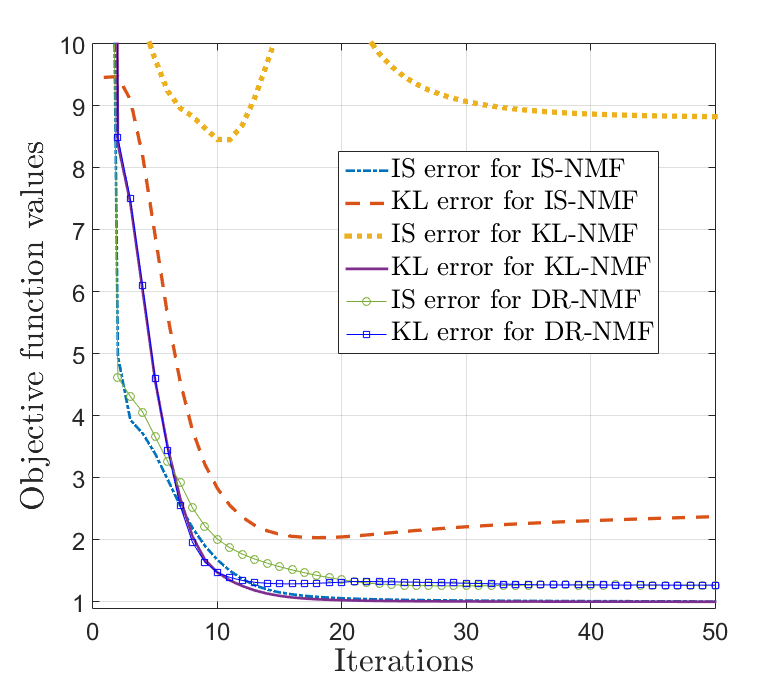}
    ~ 
    \caption{Comparative study of NMF with IS- and KL-divergences, and DR-NMF with $\Omega = \{0,1\}$ applied to the amplitude spectrogram of ``Mary had a little lamb'' with $r =4$. The figure shows the evolution of the scaled $\beta$-divergences fo the different NMF models. \label{fig:set1}}
\end{figure}
However, the three solutions generated by IS-NMF, KL-NMF and DR-NMF all give a correct separation.  
\ngi{The reason is that this recording is of good quality hence the noise is rather low.} 

\paragraph{Poisson noise} 

The second comparative study is performed on the same data set with Poisson noise added  to the input audio spectrogram following the methodology described in Section~\ref{sec:pareto}. We use $N = \epsilon  \frac{\|X\|_F}{\|\tilde{N}\|_F} \tilde{N}$ with $\epsilon = 0.6$ and $\tilde{N}=\frac{N_{\text{KL}}}{\|N_{\text{KL}}\|_F}$. Figure \ref{fig:set3} displays the rows of $H$ (that is, the activations of the notes over time) for NMF with IS- and KL-divergences, and for DR-NMF with $\Omega = \{0,1\}$ with $r=4$. 
\begin{figure}[h!]
    \centering
    ~ 
  \includegraphics[width=0.47\textwidth]{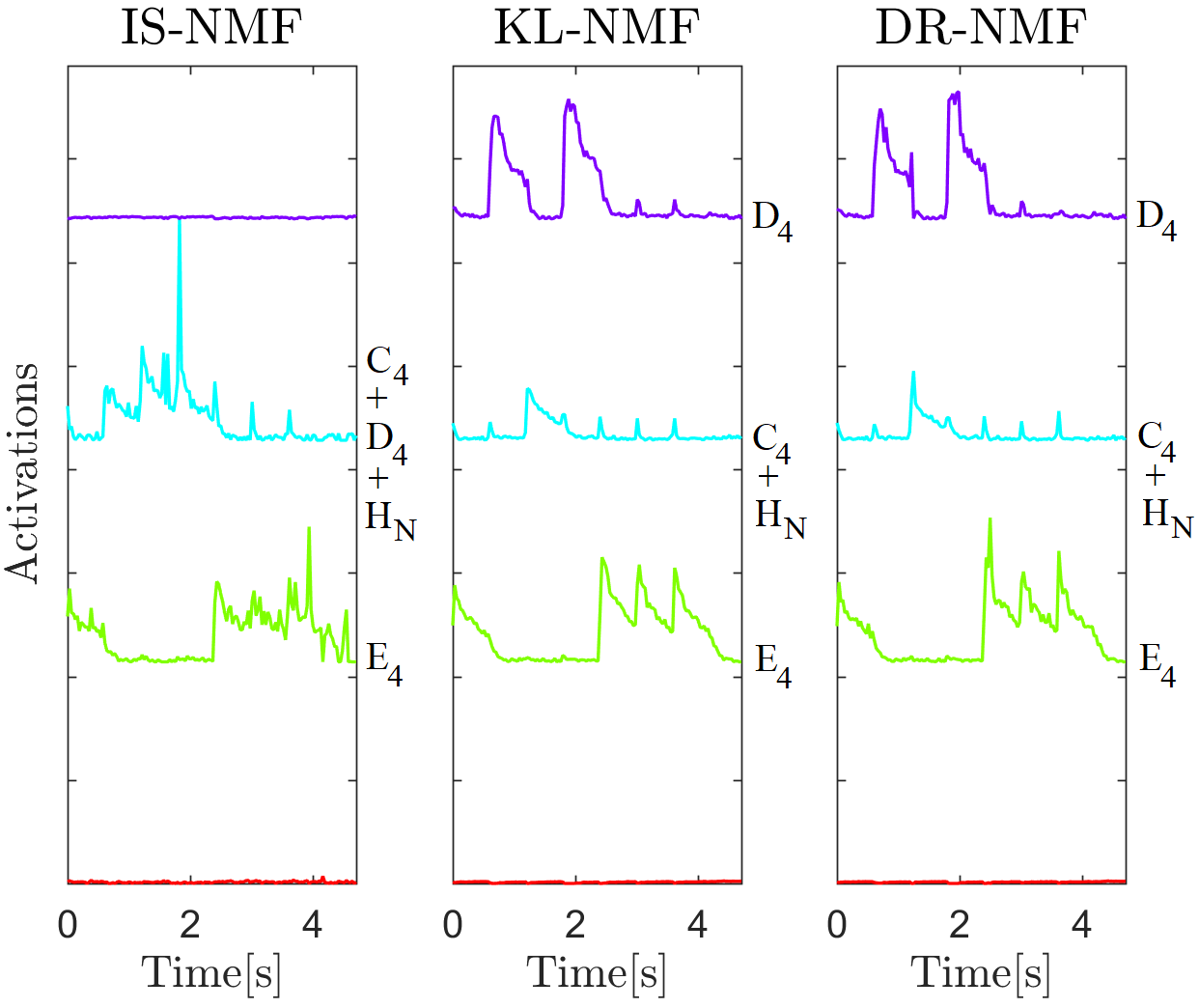}     \caption{Comparative study of NMF with IS- and KL-divergences, and DR-NMF with $\Omega = \{0,1\}$ applied to ``Mary had a little lamb'' amplitude spectrogram with $r=4$ and Poisson noise. The figure shows the activations (that is, the rows of $H$) of the recovered sources over time, \reviselast{and indicates which source it corresponds to (the notes $C_4$, $D_4$, $E_4$, and the hammer noise $H_N$)}. \label{fig:set3} }
\end{figure}

As expected with this noise model and high noise level, IS-NMF is not able to extract the three notes, while KL-NMF and DR-NMF \reviselastlast{identify them}.  
In fact, the recovered activations, that is, the rows of $H$,  correspond to the activations of the notes from the musical score shown on Figure~\ref{fig:mary_representations}: $C_4$ is activated once, $D_4$ twice and $E_4$ four times. Note that the \reviselast{hammer noise ($H_N$)} is not extracted (a source is set to zero) \reviselastlast{but is mixed with $C_4$ and to a smaller extent with $D_4$}.     
This illustrates that DR-NMF is robust to different types of noises (in this case, Poisson noise). 


\paragraph{Gamma noise} 

The third comparative study is performed on the same data set with  multiplicative Gamma noise, accordingly to the the methodology described in Section~\ref{sec:pareto}. We use $N = \epsilon  \frac{\|X\|_F}{\|\tilde{N}\|_F} \tilde{N}$ with $\epsilon = 0.4$ and $\tilde{N}=\frac{N_{\text{IS}}}{\|N_{\text{IS}}\|_F}$. For this experiment, we overestimate the number of sources present into the input spectrogram by choosing $r=5$; this allows to highlight the differences between the different NMF variants better. 
Figure~\ref{fig:set4} displays the rows of $H$ for NMF with IS- and KL-divergences, and for DR-NMF with $\Omega = \{0,1\}$. 
\begin{figure}[h!]
    \centering
    ~ 
    ~ 
        \includegraphics[width=0.47\textwidth]{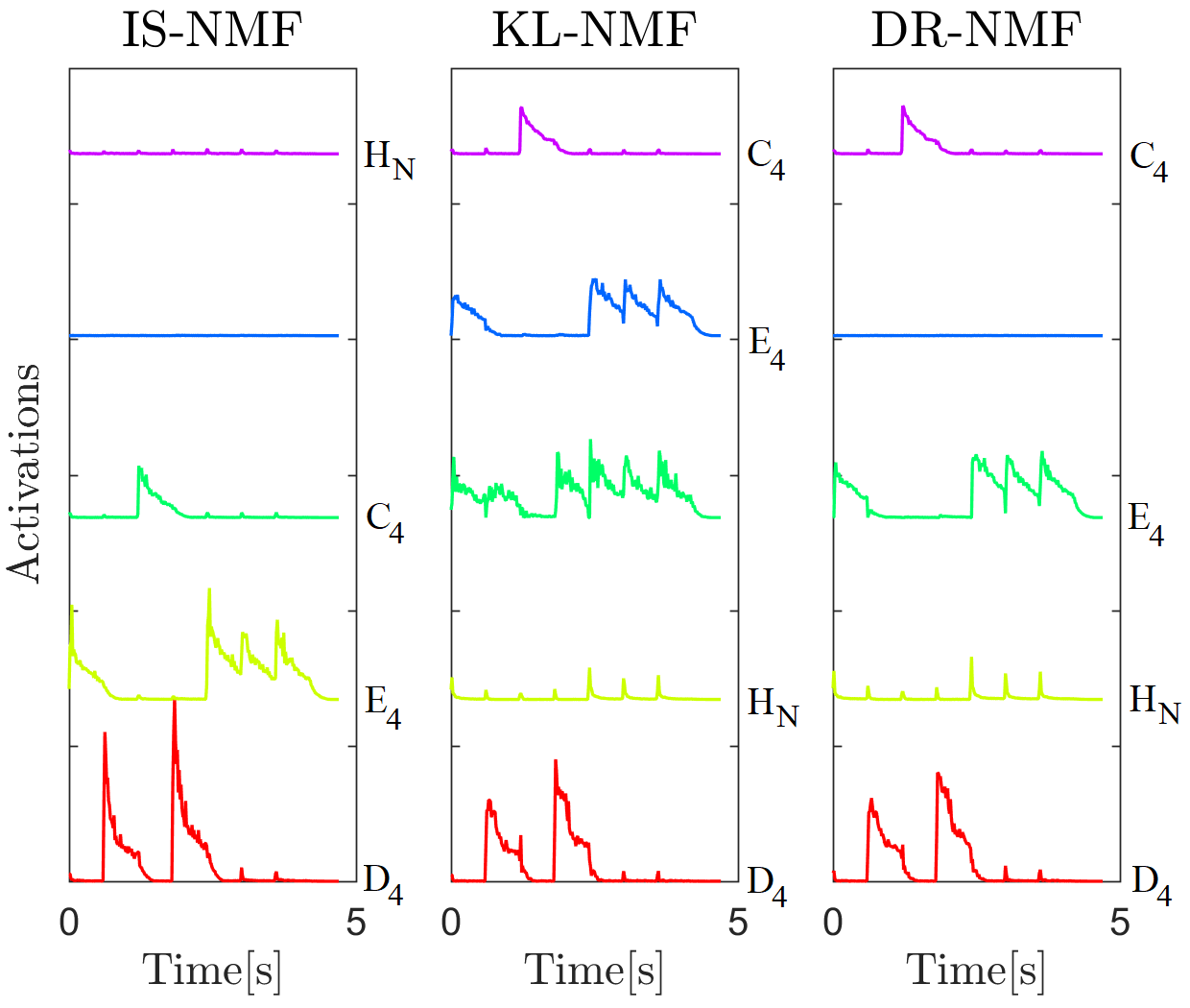} 
    \caption{Comparative study of NMF with IS- and KL-divergences, and DR-NMF with $\Omega = \{0,1\}$ applied to ``Mary had a little lamb'' amplitude spectrogram with $r =5$ and Gamma noise. The figure shows the activations (that is, the rows of $H$) of the recovered sources over time. \reviselastlast{Note that the hammer noise is also extracted with $C_4$ and $D_4$, although with a small intensity}. \label{fig:set4} }
\end{figure}

KL-NMF identifies five sources among which the third one has no physical meaning and seems to be a mixture of several notes. IS-NMF correctly identifies the three notes, the fourth estimate (the hammer) is less accurately estimated in terms of amplitude for the activations but IS-NMF is able to set to zero the fifth estimate which is appealing as it automatically remove an unnecessary component.  
DR-NMF again takes advantage from both divergences as it is able to extract the three notes correctly, the fourth estimate (the hammer) is well extracted and the fifth estimate is close to zero. This again illustrates that DR-NMF is robust to different types of noises (in this case, multiplicative Gamma noise). 


\section{Conclusion and further work} 

In this paper, we have proposed an NMF model that takes into account several data fitting terms. We then proposed to tackle this problem with a weighted-sum approach with carefully chosen weights, and designed variations of MU algorithm to minimize the corresponding objective function. We used this model to design a  DR-NMF algorithm, \color{black} inspired from the Frank-Wolfe algorithm, 
\color{black} that allows to obtain NMF solutions with low reconstruction errors with respect to several objective functions. We illustrated the effectiveness of this approach on synthetic, document and audio data sets. 
For audio data sets, DR-NMF provided particularly stunning results, being able to obtain solutions with significantly lower IS and KL errors (simultaneously), while generating meaningful solutions under different noise models or statistics. 
\color{black}
It is our hope that the proposed algorithms for DR-NMF (Algorithm~\ref{algo:drnmf}) resolve the long-standing debate~\cite{virtanen2007monaural, fevotte2009nonnegative} on whether to use IS- or KL-NMF for audio data sets. 
\color{black} 
Using DR-NMF provides a safe alternative when one is uncertain of the noise statistics of audio data sets. Indeed,  the noise statistics is rarely, if at all, known in practice. 

Possible further research include 
the design of more efficient algorithms to solve multi-objective NMF, 
the extension of our distributionally robust model to low-rank tensor decompositions, 
and the refinment of our model by adding additional penalty terms or contraints to exploit properties, such as sparsity, smoothness or minimum volume~\cite{cichocki2009nonnegative, gillis2014, fu2018nonnegative}, in the decompositions. Another challenging direction of research is to consider the DR-NMF problem with an uncountably infinite  uncertainty set $\Omega$ such as $\Omega = [0,2]$. \ngi{Finally, an important direction of research that we plan to investigate is the design of an efficient algorithm for DR-NMF with convergence guarantees; see Section~\ref{sec:algoDRNMF}.}

\section*{Acknowledgments}

We thank the reviewers and the handling editor for their insightful comments that helped us improve the paper.


%
%
%
%

%

\ifCLASSOPTIONcaptionsoff
  \newpage
\fi



%
%
%

\bibliographystyle{IEEEtran}
\bibliography{biblio} 

\end{document}